\documentclass[11pt]{article}

\usepackage[table,xcdraw]{xcolor}
\usepackage{acl}
\usepackage{booktabs}    % 用于改善表格样式
\usepackage{multirow}    % 用于多行表头
\usepackage{caption}     % 用于改善表格标题样式
\usepackage{arydshln}    % 用于支持 \hdashline
\usepackage{enumitem}
\usepackage{times}
\usepackage{latexsym}
\usepackage{amsmath}
\usepackage{amssymb}
\usepackage{subcaption}
\usepackage{array}
\usepackage{geometry}
\usepackage{xcolor}
\usepackage{pgfplots}
\usepackage{adjustbox}
\usepackage{booktabs}
\usepackage{colortbl}  % 用于表格颜色
\usepackage{xcolor}    % 用于定义颜色
\definecolor{graybg}{rgb}{0.95, 0.95, 0.95} % 定义浅灰色

\usepackage[T1]{fontenc}
\usepackage[utf8]{inputenc}

\usepackage{microtype}

\usepackage{inconsolata}

\usepackage{graphicx}
\title{Diffusion-CAM: Faithful Visual Explanations for dMLLMs}

\author{\textbf{Haomin Zuo\textsuperscript{2}},
 \textbf{Yidi Li\textsuperscript{3}}, \textbf{Luoxiao Yang\textsuperscript{4}}, \textbf{Xiaofeng Zhang\textsuperscript{1}$^\dagger$}
\\
 \textsuperscript{1}Department of Automation and Intelligent Sensing, Shanghai Jiao Tong University \\
 \textsuperscript{2}Sun Yat-sen University
 \textsuperscript{3}Northwestern University \\
  \textsuperscript{4}Technion - Israel Institute of Technology \\
    {\tt\small \{framebreak@\}sjtu.edu.cn}\quad
}

\begin{document}
\maketitle
\begin{abstract}
While diffusion Multimodal Large Language Models (dMLLMs) have recently achieved remarkable strides in multimodal generation, the development of interpretability mechanisms has lagged behind their architectural evolution. Unlike traditional autoregressive models that produce sequential activations, diffusion-based architectures generate tokens via parallel denoising, resulting in smooth, distributed activation patterns across the entire sequence. Consequently, existing Class Activation Mapping (CAM) methods, which are tailored for local, sequential dependencies, are ill-suited for interpreting these non-autoregressive behaviors. To bridge this gap, we propose \textbf{Diffusion-CAM}, the first interpretability method specifically tailored for dMLLMs. We derive raw activation maps by differentiably probing intermediate representations in the transformer backbone, accordingly capturing both latent features and their class-specific gradients. To address the inherent stochasticity of these raw signals, we incorporate four key modules to resolve spatial ambiguity and mitigate intra-image confounders and redundant token correlations. Extensive experiments demonstrate that Diffusion-CAM significantly outperforms SoTA methods in both localization accuracy and visual fidelity, establishing a new standard for understanding the parallel generation process of diffusion multimodal systems. Code is available at \url{https://github.com/ZzzzzZhhmm/Diffusion-CAM}

\begingroup
\renewcommand\thefootnote{} % 临时清空编号内容
\footnotetext{${}^\dagger$ Corresponding author}
\endgroup
\end{abstract}

\section{Introduction}
Multimodal Large Language Models (MLLMs) have fundamentally transformed Artificial Intelligence, enabling seamless cross-modal understanding and reasoning \cite{zhao2023survey}. The dominant paradigm has long centered on autoregressive architectures—such as Qwen-VL series~\cite{wang2024qwen2,bai2023qwen,qwen3technicalreport}, alongside others \cite{alayrac2022flamingo,liu2023visual,achiam2023gpt}—which generate text sequentially based on explicit attention mechanisms. However, a significant paradigm shift is underway. Recent innovations are pioneering diffusion-based architectures, exemplified by LaViDa \cite{li2025lavidaO,li2025sparse,li2025lavida,li2026lavida}, LLaDA-V \cite{you2025llada,nie2025large}, MMaDA \cite{yang2025mmada}, and Dream-VL \cite{ye2025dream2,ye2025dream}. Unlike autoregressive models that function as a ``sequential chain,'' these diffusion MLLMs employ parallel masked diffusion, conceptualizing the entire sentence simultaneously. This shift enhances both generation speed and global coherence \cite{song2019generative,song2020denoising}. But as the community pivots toward these parallel architectures, a critical question arises: \textit{Are our interpretability tools keeping pace?}
\begin{figure}[t] 
	\centering
	\includegraphics[trim={4.1cm 3cm 7.5cm 2.9cm}, clip, width=1\linewidth]{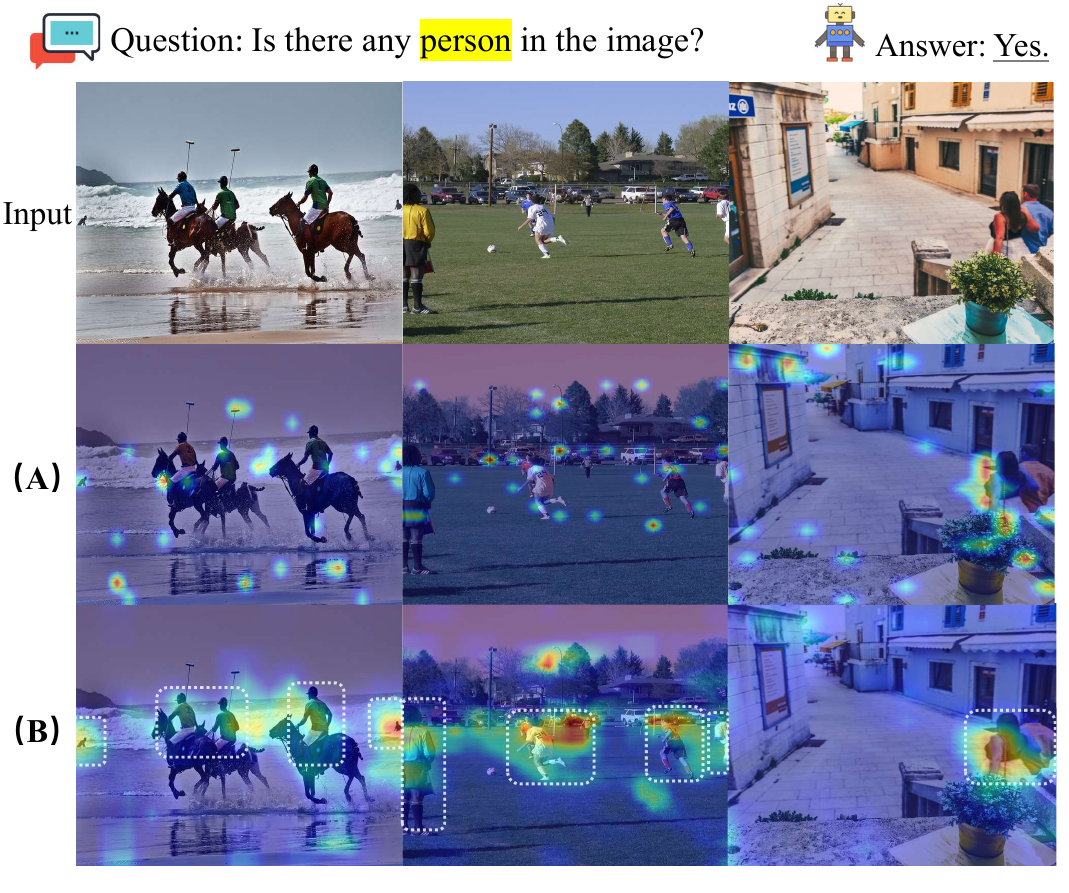}
	\caption{Visual comparison of (A) [LLaVA-CAM] and (B) [Diffusion-CAM] (ours). Our method generates more focused and clearer activation maps.}
	\label{fig:qualitative} 
\end{figure}
Understanding model decisions is a prerequisite for trustworthy AI \cite{mosbach2024insights,wang2024knowledge,pan2024dissecting}. In the autoregressive realm, visual explanation is mature, spanning from gradient-based CAMs, which provide a natural way to spatially ground model predictions by weighting feature channels with target gradients, \cite{zhou2016learning,selvaraju2017grad,jiang2021layercam,chattopadhay2018grad,li2025diffcam,chowdhury2024cape,omeiza2019smooth} to attention-based approaches \cite{abnar2020quantifying,chefer2021transformer,ferrando2024information,zhang1,zhang2,zhang3,zhang4,zhang5}. Methods like LLaVA-CAM \cite{zhang2024redundancy} and Token Activation Maps (TAM) \cite{li2025token} rely heavily on the sequential, attention-rich nature of these models to trace specific token generation. However, the very advantages of dMLLMs—\textbf{parallel generation and global planning}—pose a fundamental challenge to these frameworks. DMLLMs operate by progressively denoising a global context without explicit token-wise attention weights \cite{xu2023multimodal}. Consequently, applying traditional CAM methods results in diffuse, non-specific heatmaps (as shown in Figure \ref{fig:qualitative}), failing to disentangle the model's decision process for specific objects.

To address this limitation, we propose \textbf{Diffusion-CAM}, a gradient-based visual explanation framework tailored to the diffusion MLLMs. Unlike existing CAM-style methods built around autoregressive token dependencies, Diffusion-CAM is designed for the common dMLLM setting in which images and prompts provide fixed multimodal conditioning, while response tokens are produced through iterative parallel denoising. Our key insight is that reliable visual attribution in this setting must be extracted from \emph{structurally valid intermediate multimodal states} along the denoising trajectory, where image-grounded spatial information is still preserved and can be linked to the final prediction through gradients. Based on this principle, we first construct raw activation maps by tracing gradients from the final response back to these valid hidden features. We further find that such raw maps are intrinsically corrupted by stochastic spatial noise, diffuse background responses. To address these challenges, Diffusion-CAM introduces four complementary refinement modules, yielding more localized, faithful, and diffusion-specific visual explanations. Additionally, extensive experiments on COCO Caption \cite{chen2015microsoft} and GranDf \cite{rasheed2024glamm} datasets confirm that Diffusion-CAM achieves superior localization accuracy and background suppression compared to state-of-the-art baselines.

Our main contributions are summarized as follows:
% \begin{itemize}[leftmargin=*, noitemsep, topsep=3pt]

    % \item Identifying the Explanatory Gap: We articulate the fundamental conflict between the global, parallel nature of emerging dMLLMs and the sequential assumptions of existing explanation methods.
    % \item Diffusion-CAM Framework: We propose the first comprehensive interpretation method for dMLLMs, introducing a specialized pipeline—comprising \textbf{critical-step gradient extraction} and dedicated post-processing modules—to achieve precise localization.
    % \item Empirical Superiority: We demonstrate that Diffusion-CAM significantly outperforms state-of-the-art baselines in both localization accuracy and explanation quality across multiple benchmarks, establishing a new standard for interpreting diffusion-based multimodal models.
    \begin{itemize}
       \item We articulate the fundamental conflict between the global, parallel nature of emerging dMLLMs and the sequential assumptions of existing explanation methods.
    \item We propose the first comprehensive interpretation method for dMLLMs, introducing a specialized pipeline—comprising critical-step gradient extraction and dedicated post-processing modules—to achieve precise localization.
    \item We demonstrate that Diffusion-CAM significantly outperforms state-of-the-art baselines in both localization accuracy and explanation quality across multiple benchmarks, establishing a new standard for interpreting dMLLMs.
\end{itemize}

\section{Related Work}

\noindent\textbf{Interpretation for Autoregressive MLLMs.} Model interpretability spans mechanistic analysis~\cite{elhage2021mathematical,olsson2022context,liao-etal-2025-exploring,nam2025causal}, attribution methods~\cite{sundararajan2017axiomatic,shrikumar2017learning}, and visual explanation~\cite{selvaraju2017grad,zhou2016learning}. 
With the prominence of autoregressive MLLMs like LLaVA~\cite{liu2023visual,liu2024improved}, Gemini~\cite{team2023gemini}, and others~\cite{chen2024internvl,bai2023qwen}, visual explanation has become vital for tracing token generation. 
Current techniques range from gradient-based CAMs~\cite{selvaraju2017grad,jiang2021layercam,smilkov2017smoothgrad} and attention mechanisms~\cite{chefer2021transformer,abnar2020quantifying,montavon2019layer} to perturbation methods like LIME and SHAP~\cite{ribeiro2016should,lundberg2017unified}. 
While these approaches inherently rely on autoregressive dependencies~\cite{vaswani2017attention}, rendering them unsuitable for non-sequential architectures.

\noindent\textbf{Challenges in Diffusion Architectures.}
The shift from AR MLLMs to dMLLMs~\cite{you2025llada,ye2025dream2,li2025lavidaO,li2025sparse,li2025lavida,li2026lavida} creates a different interpretability setting. 
Instead of left-to-right decoding, dMLLMs generate responses by iterative masked denoising~\cite{ho2020denoising,song2020denoising} under fixed multimodal conditioning, so the token-level causal structure assumed by conventional CAM-style methods no longer directly applies. 
As a result, visual attribution must be extracted from intermediate multimodal states that still preserve image-grounded spatial structure during denoising. 
Meanwhile, existing diffusion-interpretability methods for text-to-image generation, such as DAAM~\cite{tang2023daam}, don't directly transfer to multimodal reasoning, and diffusion-based global refinement can further amplify noise from feature redundancy, background dispersion, and unstable visual representations~\cite{darcet2023vision,he2016deep,li2025closer,balasubramanian2024decomposing}. 
Our work is motivated by these diffusion-specific challenges.

\section{Method}
In this section, we first explain how we utilize the common structural and principleal characteristics of dMLLMs to adapt the CAM method to them. Secondly, we expound four modules in Diffusion-CAM specifically designed to improve the clarity and effectiveness of activation maps. Finally, we introduce four metrics designed to enable fine-grained evaluations of the explanations.
\subsection{CAM Adaptation for dMLLMs}
\begin{figure*}[htbp]
	\centering
	\includegraphics[trim={1cm 6.25cm 0.8cm 6.1cm}, clip,width=1\textwidth]{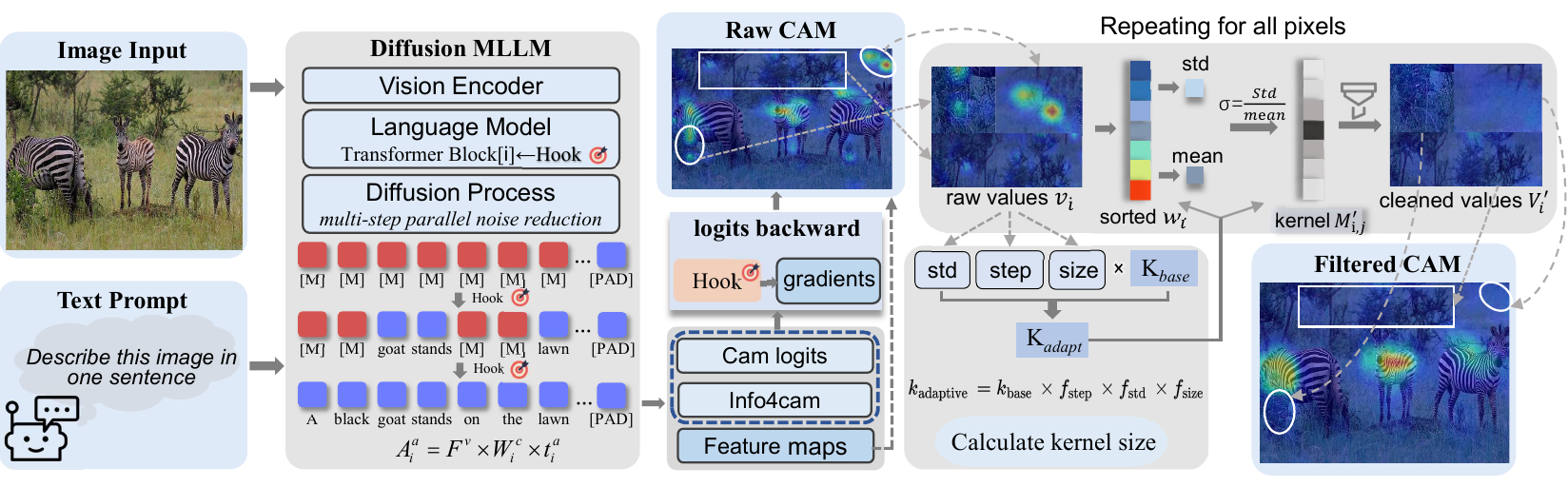}
	\caption{\textbf{Illustration of the Raw CAM generation and Adaptive Denoising.} \textbf{(Left) Raw CAM Generation:} We capture intermediate visual features and gradients from the LaViDa backbone during the denoising process via hook registration. \textbf{(Right) Adaptive Kernel Denoising:} A dynamic kernel $\mathbf{K}_{adapt}$, calibrated by feature statistics (e.g., $\sigma$), drives a Rank Gaussian Filter to suppress stochastic noise and architectural artifacts, transforming noisy activations into precise, semantically coherent heatmaps.}
	\label{fig:CAM_generation}
\end{figure*}
Conventional gradient-based visual explanation techniques~\cite{selvaraju2017grad,zhang2024redundancy} were primarily developed for autoregressive vision-language models~\cite{liu2023visual,bai2023qwen}. In autoregressive architectures, the model predicts tokens sequentially, with each token $t_i$ conditioned on all preceding tokens: $p(t_i \mid t_{<i}, \mathbf{I})$. This left-to-right factorization yields a clear gradient path from the target output to the visual features, making CAM computation relatively direct.

Recent dMLLMs~\cite{li2025lavida,you2025llada,yang2025mmada,li2025lavidaO,li2026lavida,li2025sparse}, despite differing in task scope and implementation details, share a common generative mechanism: image features and textual prompts serve as fixed multimodal conditioning, while the response tokens are generated through \emph{iterative masked denoising} rather than next-token prediction. This changes the attribution problem fundamentally. Instead of tracing gradients through a single autoregressive decoding path, CAM for dMLLMs must identify a denoising step whose intermediate hidden states still preserve the image-conditioned spatial structure required for visual grounding. Accordingly, we adapt CAM to this shared conditional masked-diffusion interface rather than to any model-specific decoding heuristic.

Formally, for autoregressive models, CAM computation follows:
\vspace{-0.4 cm}
\begin{equation}
    G_k = \frac{\partial y^c}{\partial A_k}
\end{equation}

For diffusion models, we instead compute gradients from the final response score to image-grounded features at a \emph{valid conditioning step}:
\vspace{-0.1 cm}
\begin{equation}
    G_c^{(s)} = \frac{\partial \mathcal{L}_{\text{final}}}{\partial A_c^{(s)}}, 
    \qquad
    \mathcal{L}_{\text{final}} = \sum_{t \in \mathcal{T}} \mathbf{z}_{\text{final}}[t]
\end{equation}
where $\mathcal{T}$ denotes the selected answer-token indices, and $A_c^{(s)}$ represents the $c$-th feature channel at denoising step $s$. In our implementation, the attribution step is \emph{not hard-coded}. Instead, we select it using a dynamic feasibility criterion: a denoising step is valid only if the hooked hidden-state sequence still contains the full image-token span required for image-region extraction. For example, under LaViDa's Prefix-DLM inference~\cite{li2025lavida}, this criterion is satisfied only at the earliest conditioning step $s=0$~\ref{app:shared_dmlm_cam}; for other dMLLMs, the same rule naturally extends to whichever step(s) remain structurally valid.

We introduce three adaptations to make CAM compatible with this setting.

\noindent\textbf{(1) Model-aware feature extraction.}
We register a forward hook at an intermediate transformer block and retain its gradient during backpropagation from the final prediction score. This layer is selected from a layer sweep as a stable trade-off between spatial grounding and multimodal semantic integration. Rather than assuming a fixed timestep a priori, we extract features only from denoising steps that satisfy the above feasibility condition.

\noindent\textbf{(2) Dynamic image-span localization.}
Because dMLLM hidden states contain mixed multimodal tokens, image features must be localized dynamically rather than assumed to occupy fixed positions. We store multimodal packing metadata in \texttt{info4cam}, from which we parse the image-token boundaries. Given the full hidden sequence $\mathbf{F} \in \mathbb{R}^{L \times D}$, we extract image features as:
\begin{equation}
\begin{aligned}
    \mathbf{A}_{\text{img}} &= \mathbf{F}[\mathcal{I}_{\text{img}}] \in \mathbb{R}^{(H \times W) \times D}, \\
    \mathcal{I}_{\text{img}} &= \{ i \in \mathbb{N} : N_{\text{base}} \le i < N_{\text{base}} + H \times W \},
\end{aligned}
\end{equation}
where $N_{\text{base}}$ is the base text/prompt offset recovered from \texttt{info4cam}. The extracted token sequence is then reshaped to $\mathbb{R}^{D \times H \times W}$ to form the spatial feature map.

\noindent\textbf{(3) Diffusion-CAM generation.}
On the valid image-region feature map, we apply a Grad-CAM-style aggregation. Specifically, we spatially average the gradients to obtain channel weights
\vspace{-0.3 cm}
\begin{equation}
\begin{aligned}
    w_c &= \frac{1}{H \times W} \sum_{i,j} G_c^{(s)}(i,j), \\
    M_{\text{base}} &= \mathrm{ReLU}\left( \sum_c w_c \cdot A_c^{(s)} \right).
\end{aligned}
\end{equation}
This yields the baseline CAM before the subsequent refinement modules. In this way, the proposed adaptation is tied to the shared masked-denoising structure of dMLLMs, rather than to the implementation details of a single model.
\subsection{Four Modules within Diffusion-CAM}
\subsubsection{Adaptive Kernel Denoising}
Transformer-based multimodal models exhibit high-frequency architectural artifacts in activation maps due to the discrete nature of self-attention. To mitigate this, we introduce a \textbf{ Adaptive Kernel Denoising} module. This approach synergizes the robustness of order statistics with Gaussian weighting, dynamically calibrating filtering parameters based on diffusion properties.

\noindent\textbf{Dynamic Kernel Determination.}
Standard fixed-size filters fail to accommodate varying noise profiles across different timesteps. As shown in the right of Figure~\ref{fig:CAM_generation}, we propose a dynamic strategy scaling the receptive field $k_{\text{adaptive}}$ via three governing factors:
\begin{equation}
    k_{\text{adaptive}} = \lfloor k_{\text{base}} \cdot \mathcal{F}_{\text{step}}(S) \cdot \mathcal{F}_{\text{std}}(\sigma_{\mathbf{M}}) \cdot \mathcal{F}_{\text{size}}(H) \rfloor_{\text{odd}}
\end{equation}
where $\lfloor \cdot \rfloor_{\text{odd}}$ yields the nearest odd integer. $\mathcal{F}_{\text{step}}$ scales with denoising steps $S$, as longer trajectories produce diffuse semantic patterns requiring larger aggregation windows. $\mathcal{F}_{\text{std}}$ reacts to spatial variance $\sigma_{\mathbf{M}}$, widening the kernel to suppress aggressive peaks in high-noise scenarios, while $\mathcal{F}_{\text{size}}$ adjusts for resolution $H$ to ensure scale invariance.

\noindent\textbf{Rank-Weighted Gaussian Filtering.}
To remove artifacts without blurring semantic boundaries, we employ a Rank-Weighted Gaussian Filter. Unlike spatial convolution, we sort activation values within a local window $\Omega_{i,j}$ to form an ordered set $\mathcal{V}^{\text{sorted}}$. The filtered value $M'_{i,j}$ is computed as:
\vspace{-0.3cm}
\begin{equation}
    M'_{i,j} = \sum_{n=1}^{k^2} v_n \cdot \mathcal{G}_n(\mu, \sigma_{rank})
\end{equation}
Here, $\mathcal{G}_n$ applies normalized Gaussian weights to the \textbf{rank index} $n$ rather than spatial distance. This mechanism effectively suppresses outliers occupying extreme ranks while preserving the dominant semantic signal, offering superior robustness over standard linear filtering.

\begin{figure*}[htbp]
	\centering
	\includegraphics[trim={0.15cm 4.6cm 0.2cm 4.4cm}, clip,width=1.02\textwidth]{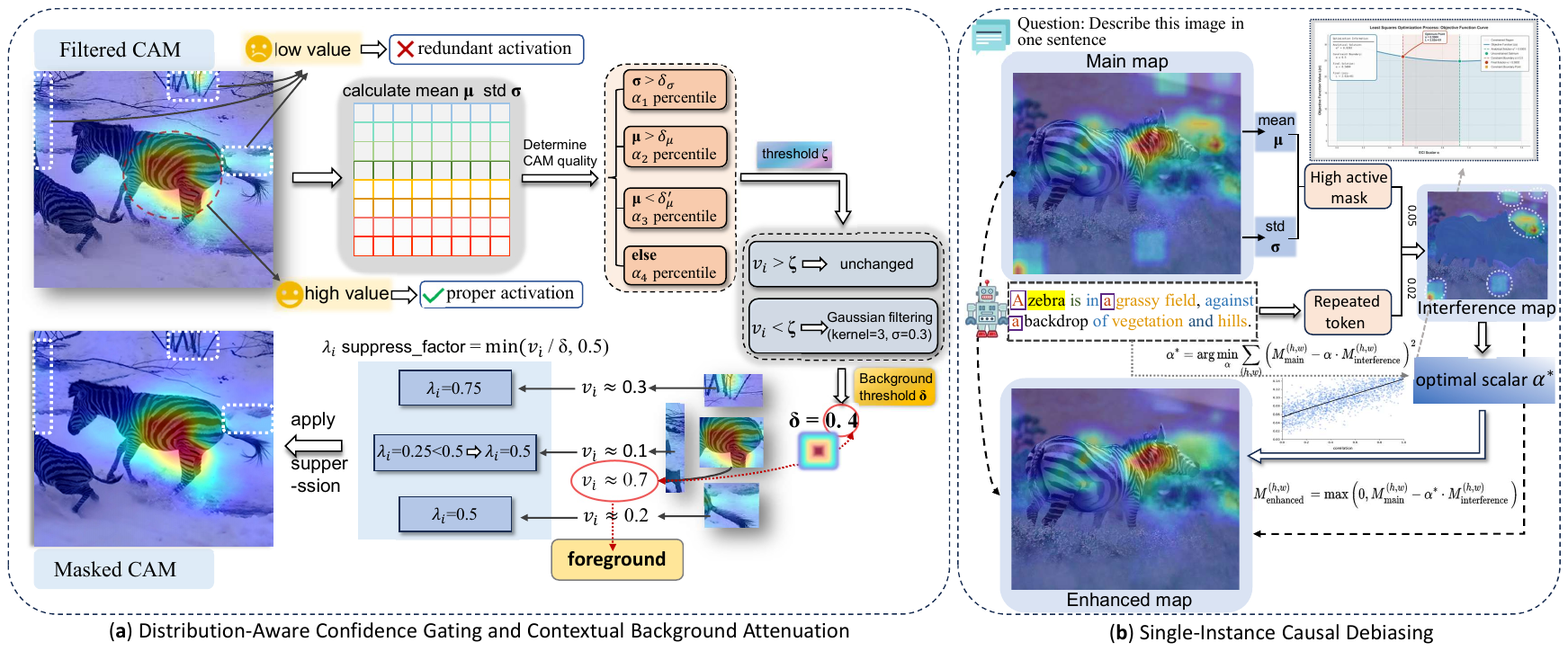}
	\caption{Illustration of the other three modules of Diffusion-CAM. \textbf{Distribution-Aware Confidence Gating:} We calculate an adaptive threshold and process regions differently. \textbf{Contextual Background Attenuation:} identifies the foreground and background and calculates suppression factors. \textbf{Single-Instance Causal Debiasing:} uses repeated tokens and abnormally high activation mask to make heatmap clearer.}
	\label{fig:DACD SICD}
\end{figure*}

\subsubsection{Distribution-Aware Confidence Gating}
Different diffusion steps and image contents manifest varying statistical properties in activation maps. Uniform processing strategies often induce artifacts in high-variance scenarios while failing to denoise low-variance ones effectively. To mitigate this stochasticity, we propose \textbf{Distribution-Aware Confidence Gating (DACG)}, as illustrated in Figure~\ref{fig:DACD SICD}(a).

Instead of rigid thresholding, DACG dynamically calibrates the confidence boundary $\tau_{\text{conf}}$ by analyzing the global activation statistics (mean $\mu_{\mathbf{M}}$ and standard deviation $\sigma_{\mathbf{M}}$). We formulate a dynamic mapping function $\mathcal{F}(\cdot)$ that determines the optimal quantile $\alpha$ for filtering:
\vspace{-0.1cm}
\begin{equation}
\begin{aligned}
    \tau_{\text{conf}} &= \text{Quantile}(\mathbf{M}, \alpha), \\
    \alpha &= \mathcal{F}(\mu_{\mathbf{M}}, \sigma_{\mathbf{M}})
\end{aligned}
\label{eq:1}
\end{equation}
Specifically, $\alpha$ adaptively shifts between stricter bounds (e.g., 90th percentile) for high-variance distributions ($\sigma_{\mathbf{M}} > \delta_{\sigma}$) to suppress noise, and relaxed bounds (e.g., 75th percentile) for high-brightness maps ($\mu_{\mathbf{M}} > \delta_{\mu}$) to preserve signals. This statistical alignment ensures robustness across diverse generation scenarios.

Crucially, we implement a \textbf{selective processing mechanism}. Rather than applying uniform denoising, we partition the activation map into high-confidence ($\mathbf{M}_{hc}$) and low-confidence ($\mathbf{M}_{lc}$) zones based on $\tau_{\text{conf}}$:
\begin{equation}
\begin{aligned}
     \mathbf{M}_{hc} = \{(i,j) \mid M(i,j) > \tau_{\text{conf}}\}\\
    \mathbf{M}_{lc} = \{(i,j) \mid M(i,j) \le \tau_{\text{conf}}\}
\end{aligned}
\label{eq:2}
\end{equation}
Then we apply mild Gaussian Denoising (kernel=3, $\sigma$=0.3) to $\mathbf{M}_{lc}$ to reduce salt-and-pepper noises, while $\mathbf{M}_{hc}$ retains its raw structural integrity. This dual-path strategy effectively disentangles semantic structures from background noise without over-smoothing salient regions.

\subsubsection{Contextual Background Attenuation}

In dMLLMs, background regions often accumulate residual signals from multiple denoising steps, leading to boundary blurring. To address this, as shown in Figure~\ref{fig:DACD SICD}(a), we introduce \textbf{Contextual Background Attenuation (CBA)}, which leverages a multi-scale statistical ensemble to define a robust separation boundary.

We construct a composite background threshold $\tau_{\text{bg}}$ by aggregating complementary statistical descriptors:
\vspace{-0.4 cm}
\begin{equation}
\tau_{\text{bg}} = \sum_{k} w_k \cdot \mathcal{S}_k(\mathbf{M})
\end{equation}
where $\mathcal{S}(\mathbf{M})$ represents a vector of statistical metrics including the median of non-zero activations, global distribution quantiles, and peak activation references. For example, let \(\mathbf{M} \in \mathbb{R}_{\ge 0}^{H\times W}\) be the normalized CAM map, and \(\mathbf{M}_{>0}=\{\mathbf{M}_{i,j}\mid \mathbf{M}_{i,j}>0\}\). We define the descriptor vector:
\begin{equation}
\mathcal{S}(\mathbf{M})=[S_1,S_2,S_3,S_4],
\end{equation}
with
$\mathcal{S}_1=\mathrm{median}(\mathbf{M}_{>0}), \mathcal{S}_2=Q_{0.60}(\mathbf{M}), \mathcal{S}_3=\mathbb{E}[\mathbf{M}_{>0}], \mathcal{S}_4=\max(\mathbf{M}).$

The weights $w_k$ balance the contribution of conservative estimation and strong signal anchoring, ensuring the threshold adapts to the image's specific contrast profile.

For identified background regions $\mathcal{B} = \{(i,j) \mid \mathbf{M}_{i,j} < \tau_{\text{bg}}\}$, we employ a \textbf{progressive soft-attenuation} function instead of hard truncation:
\begin{equation}
\mathbf{M}'_{i,j} = \mathbf{M}_{i,j} \cdot \max\left(\gamma, \frac{\mathbf{M}_{i,j}}{\tau_{\text{bg}}}\right), \quad \forall (i,j) \in \mathcal{B}
\end{equation}
Here, $\gamma$ (set to 0.5) serves as a retention lower bound. This soft-masking approach creates a smooth gradient transition, effectively suppressing background noise while preventing the introduction of artificial hard edges common in power-law suppression, thereby enhancing the foreground-background contrast ratio naturally.

\subsubsection{Single-Instance Causal Debiasing}
Interference refers to activation in the heatmap that is weakly related to the intended visual evidence,such as from tokens unrelated to the key answers and from prompt tokens. Standard ECI~\cite{li2025token} relies on cross-image frequency analysis to identify confounders, rendering it inapplicable to the single-instance inference paradigm of dMLLMs. To bridge this gap, we propose \textbf{Single-Instance Causal Debiasing (SICD)}, grounded in the \textit{Hypothesis of Linguistic Economy}.

We observe that human descriptions typically employ referential grouping for semantic entities (e.g., ``two sheep'' rather than ``a sheep... a sheep''), whereas repetition is predominantly exhibited by \textbf{syntactic function words} (e.g., ``a'', ``the''). Furthermore, the phenomenon of language economy has also been verified in some studies~\cite{meister2021revisiting,gwak2025revisiting,hao2025uniform,jaeger2006speakers}. For instance, in daily communication, humans have many ways to express the same meaning. Rational speakers will subconsciously follow the Uniform Information Density (UID) assumption when organizing sentences, and try to avoid \textit{peaks} and \textit{valleys} of information density in the sentences, such as omitting the relative pronoun ``that''. In the VQA visualization activation maps for MLLMs, unlike semantic nouns that ground specific objects, these high-frequency functional tokens generate diffuse, non-specific activations that act as background noise. Consequently, as shown in Figure~\ref{fig:DACD SICD}(b), we reformulate interference $\mathcal{I}$ by targeting these syntactic redundancies alongside statistical anomalies:
\vspace{-0.2 cm}
\begin{equation}
\mathcal{I} = \omega_{\text{rep}} \cdot \mathbf{M}_{\text{rep}} + \omega_{\text{out}} \cdot \mathbf{M}_{\text{out}}
\end{equation}
where $\mathbf{M}_{\text{rep}}$ captures the diffuse activations from repeated functional tokens, and $\mathbf{M}_{\text{out}}$ isolates spatial outliers exceeding statistical bounds (e.g., $\mu_{\mathbf{M}} + 2\sigma_{\mathbf{M}}$). $\omega_{\text{rep}}$ and $\omega_{\text{out}}$ are balancing coefficients.

To determine the optimal intervention strength $\lambda^*$, we employ a constrained least-squares objective:
\vspace{-0.3 cm}
\begin{equation}
\lambda^* = \arg\min_{\lambda \in [0, \lambda_{\max}]} \|\mathbf{M} - \lambda \cdot \mathcal{I}\|_F^2
\end{equation}

The final corrected map is obtained via $\mathbf{M}_{\text{clean}} = \text{ReLU}(\mathbf{M} - \lambda^* \cdot \mathcal{I})$. This module operates as a conditional gate, triggering only when the map exhibits pathological traits (high variance/skewness), thus preserving the integrity of non-repetitive, semantically rich descriptions.
\subsection{Evaluation Metrics}

We propose a three-metric evaluation framework tailored for diffusion models' smooth, gradually distributed activations~\cite{li2025token}.

\noindent\textbf{Target Localization (Obj-IoU):} Measures spatial overlap between predicted regions $\mathcal{P}$ (obtained via Otsu thresholding) and ground truth $\mathcal{G}_c$ for object class $c$:
\vspace{-0.4 cm}
\begin{equation}
\text{Obj-IoU} = \max_{c} \frac{|\mathcal{P} \cap \mathcal{G}_c|}{|\mathcal{P} \cup \mathcal{G}_c|}
\end{equation}

\noindent\textbf{Foreground-Background Contrast:} Quantifies background suppression effectiveness, where $M_{\text{fg}}$ and $M_{\text{bg}}$ denote mean activations within and outside ground truth regions:
\vspace{-0.2 cm}
\begin{equation}
R_{\text{c}} = \frac{\mathbb{E}[M_{\text{fg}}]}{\mathbb{E}[M_{\text{bg}}] + \epsilon}
\end{equation}

\noindent\textbf{Target Activation Concentration:} Measures activation focus on target regions, where $\mathcal{G}$ represents ground truth pixel locations:
\vspace{-0.1 cm}
\begin{equation}
C_{\text{t}} = \frac{\sum_{(i,j) \in \mathcal{G}} M(i,j)}{\sum_{(i,j)} M(i,j) + \epsilon}
\end{equation}

\noindent\textbf{F3-Score:} We employ harmonic mean to ensure balanced performance, as it is more sensitive to lower values and penalizes imbalanced development:
\vspace{-0.2 cm}
\begin{equation}
\text{F3-Score} = 3 \Big/ \left( \tfrac{1}{\text{Obj-IoU}} + \tfrac{1}{\min\!\left(\frac{R_{\text{c}}}{20},\, 1\right)} + \tfrac{1}{C_{\text{t}}} \right)
\end{equation}

\begin{table*}[!t]
\centering
\small
\renewcommand{\arraystretch}{1.4} % 保持舒适的行高
\setlength{\tabcolsep}{4.5pt}        % 稍微增加列宽，让数字呼吸

\begin{tabular}{@{}lcccccccc@{}} 
\toprule
 & \multicolumn{4}{c}{\textbf{COCO Caption}} & \multicolumn{4}{c}{\textbf{GranDf}} \\
\cmidrule(r){2-5} \cmidrule(l){6-9}
\textbf{Method} & \textbf{Obj-IoU(\%)} & \textbf{Contrast} & \textbf{Concen.} & \textbf{F3-Score} & \textbf{Obj-IoU(\%)} & \textbf{Contrast} & \textbf{Concen.} & \textbf{F3-Score} \\
\midrule

% --- Baselines ---
% 逻辑：找出每列中 Baseline (前三行) 的最大值，加上 \underline{}

LLaVA-CAM & \underline{20.02} & 2.18$\times$ & 41.22 & \underline{18.08} & 18.16 & 1.41$\times$ & 70.80 & 14.22 \\

Grad-CAM & 19.93 & 2.04$\times$ & \underline{41.91} & 17.43 & 17.82 & \underline{1.48$\times$} & \underline{75.42} & \underline{14.67} \\

TAM & 15.21 & \underline{2.51$\times$} & 40.10 & 17.61 & \underline{20.39} & 1.36$\times$ & 67.31 & 14.23 \\

% --- 分隔线或间距 ---
% \addlinespace 是 booktabs 提供的神器，比 \midrule 更轻量，适合做视觉分组

\noalign{\smallskip} \hdashline \noalign{\smallskip}
% --- Ours (Best results in Bold) ---
\textbf{Diffusion-CAM (Ours)} & \textbf{30.10} & \textbf{2.58$\times$} & \textbf{51.41} & \textbf{23.04} & \textbf{28.41} & \textbf{2.02$\times$} & \textbf{86.14} & \textbf{20.53} \\
\bottomrule
\end{tabular}
\caption{Comprehensive comparison with state-of-the-art methods on COCO Caption~\cite{chen2015microsoft} and GranDf datasets~\cite{rasheed2024glamm}, and the best baseline results are \underline{underlined}. The F3-Score (\%) reflects the overall performance, calculated as the harmonic mean of the three core metrics.}
\label{tab:comparison}
\end{table*}
\section{Experiments}
\subsection{Experimental Setup}

\noindent \textbf{Datasets}
We evaluate our approach on datasets featuring both textual descriptions and pixel-level annotations. Our primary benchmark is the COCO Caption dataset (40,504 images) \cite{chen2015microsoft}. Since interpretability methods only require inference, we directly utilize the validation set. All images include segmentation masks for quantitative evaluation. We additionally report results on the GranDf dataset (1,000 images) \cite{rasheed2024glamm}, which provides fine-grained object localization annotations. Both datasets employ manually annotated masks to ensure reliability.
\begin{table}[!t]
    \centering
    \small
    \renewcommand{\arraystretch}{1.35}
    \setlength{\tabcolsep}{3.5pt}
    %---------------------------------------------------------
    % 表格 1：Ablation Study
    %---------------------------------------------------------
    % 注意：先写表格内容，Caption放在表格下方（CVPR标准）
    % 如果你的模板要求Caption在上方，把 \caption 移到 \begin{tabular} 之前
    \begin{tabular}{@{}lcccc@{}}
    
    \toprule
    \textbf{Config.} & \textbf{Obj-IoU(\%)} & \textbf{Contrast} & \textbf{Concen.} & \textbf{F3-Score} \\
    \midrule
    Baseline & 20.12 & 2.19× & 41.42 & 18.16 \\
    + Denoising & 25.85 & 2.28× & 44.01 & 20.16 \\
    + Gating & 21.22 & 2.20× & 47.94 & 18.88 \\
    + Attenuation & 21.42 & 2.41× & 42.63 & 19.59 \\
    + Debiasing & 24.11 & 2.29× & 44.31 & 19.82 \\
    \midrule
    \textbf{Full} & \textbf{30.10} & \textbf{2.58×} & \textbf{51.44} & \textbf{23.04} \\
    \bottomrule
    \end{tabular}
    \caption{Ablation study on COCO Caption~\cite{chen2015microsoft}. Each row toggles one module on top of the baseline.}
    \label{tab:ablation}
    
    %---------------------------------------------------------
    % 关键控制点：调整这个数值来改变两个表格之间的间隙
    %---------------------------------------------------------
    \vspace{0.2cm} 
    
    %---------------------------------------------------------
    % 表格 2：Layer Comparison
    %---------------------------------------------------------
    \begin{tabular}{@{}lcccc@{}}
    \toprule
    \textbf{Method} & \textbf{Layer} & \textbf{Obj-IoU(\%)} & \textbf{Contrast} & \textbf{F3-Score} \\
    \midrule
    \multirow{4}{*}{LLaVA-CAM} & 10 & 20.02 & 2.18× & 18.07 \\
     & 20 & 13.01 & 1.44× & 11.51\\
     & 25 & 16.09 & 1.19× & 10.56\\
     & 29 & 17.16 & 1.00× & 10.04\\
    \midrule
    \multirow{4}{*}{Diffusion-CAM} & 10 & 30.10 & 2.58× & 23.04 \\
     & 20 & 29.62 & 1.58× & 15.44 \\
     & 25 & 22.37 & 1.22× & 12.23 \\
     & 29 & 19.08 & 1.03× & 10.86 \\
    \bottomrule
    \end{tabular}
    \caption{Performance comparison of Diffusion-CAM and LLaVA-CAM across different target layers on COCO Caption~\cite{chen2015microsoft}.}
    \label{tab:applicability}
\end{table}

\noindent \textbf {Implementation Details}
Our experiments were conducted on an H20 GPU. The baseline LLaVA-CAM \cite{zhang2024redundancy} employs a fixed threshold of 0.4 for feature selection, while our approach omits this step. The hyperparameters $\delta_{\sigma}$, $\delta_{\mu}$,$\delta'_{\sigma}$ in \textit{DACG} are set to 0.22, 0.35 and 0.25, which are derived from empirical observation of the activation distribution. For IoU computation and performance evaluation, we tag each word’s part of speech using the pos\_tag function from the NLTK Python package and employ standard Otsu threshold binarization. Furthermore, our results are the average values obtained from multiple experiments.

\subsection{Quantitative Results}
\noindent \textbf{Ablation Studies and Effectiveness Analysis}
Table~\ref{tab:ablation} reports the results of COCO Caption \cite{chen2015microsoft} ablation experiments with only one Diffusion-CAM module activated on the baseline model. Specifically, the \textit{Adaptive Kernel Denoising} alone lifts F3-Score by 2.0\% through suppressing salt-and-pepper noise; \textit{Distribution-Aware Confidence Gating} raises concentration from 41.42\% to 47.94\% through selective refinement and the \textit{Contextual Background Attenuation} increases foreground–background contrast from 2.19× to 2.41×, while preserving target integrity. Using \textit{Causal Debiasing} is enabled the Obj-IoU reaches 24.11\%. Results illustrate each module can make different improvements to the three metrics. Although evaluated independently, the modules are complementary: when all are activated simultaneously the F3-Score rises by 4.88\%.
\begin{figure*}[htbp]
	\centering
	\includegraphics[trim={0.68cm 5cm 1cm 3.7cm}, clip, width=1.01\textwidth]{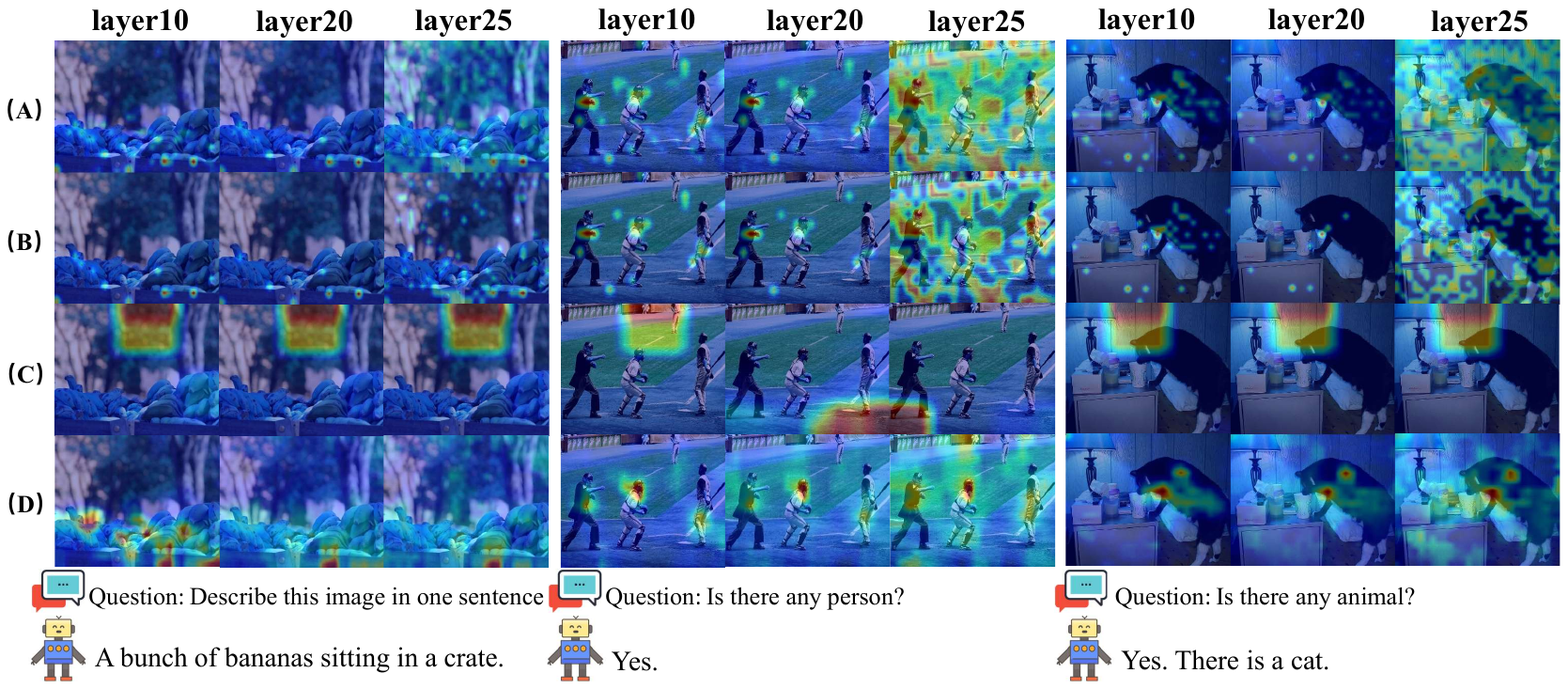}
	\caption{ 
    Visual comparison between our method (D) and SoTA approaches including (A) [Grad-CAM], (B) [LLaVA-CAM], and (C) [TAM] at different Transformer layers. 
    Diffusion-CAM produces more precise activation maps compared to the baselines, exhibiting robust localization ability across all visualized layers.
    }
	\label{fig:layer_comparison}
\end{figure*}

\noindent \textbf{Comparison with SoTA Methods}
Table~\ref{tab:comparison} presents comprehensive comparisons between Diffusion-CAM and existing methods: LLaVA-CAM \cite{zhang2024redundancy}, TAM \cite{li2025token} and Grad-CAM \cite{selvaraju2017grad}. Our approach surpasses all competitors in all four metrics. 

The comparative analysis demonstrates the superiority of our method across different datasets and evaluation metrics. On COCO Caption dataset \cite{chen2015microsoft}, Diffusion-CAM achieves an F3-Score of 23.04\%, representing a 4.96\% improvement over the strongest baseline LLaVA-CAM \cite{zhang2024redundancy}. The 30.1\% Obj-IoU indicates exceptional target localization capability, while the 2.58× contrast ratio and 51.4\% concentration validate effective background suppression and activation focusing. On the more challenging GranDf~\cite{rasheed2024glamm} benchmark, our method maintains superior performance with an F3-Score of 20.53\%, exceeding Grad-CAM \cite{selvaraju2017grad} by 5.86\%. These results confirm the robustness and generalization capability of our method across varying dataset complexities.

\noindent \textbf{Applicability and Scalability}
Diffusion-CAM exhibits broad applicability through its design leveraging core properties of diffusion architectures, making it applicable to various dMLLMs. Notably, as shown in Table~\ref{tab:applicability}, our method outperforms method LLaVA-CAM \cite{zhang2024redundancy}(which gets the highest score on layer 10 in the SoTA methods) across different target layers. It demonstrates the applicability of our method to different transformer layers. Simultaneously, the framework demonstrates excellent scalability through four independent modules that can be selectively applied to meet different scenario requirements.

\subsection{Qualitative Results}
\begin{figure*}[htbp]
	\centering
\includegraphics[trim={3.2cm 5cm 1.5cm 5.3cm}, clip, width=1.0\linewidth]{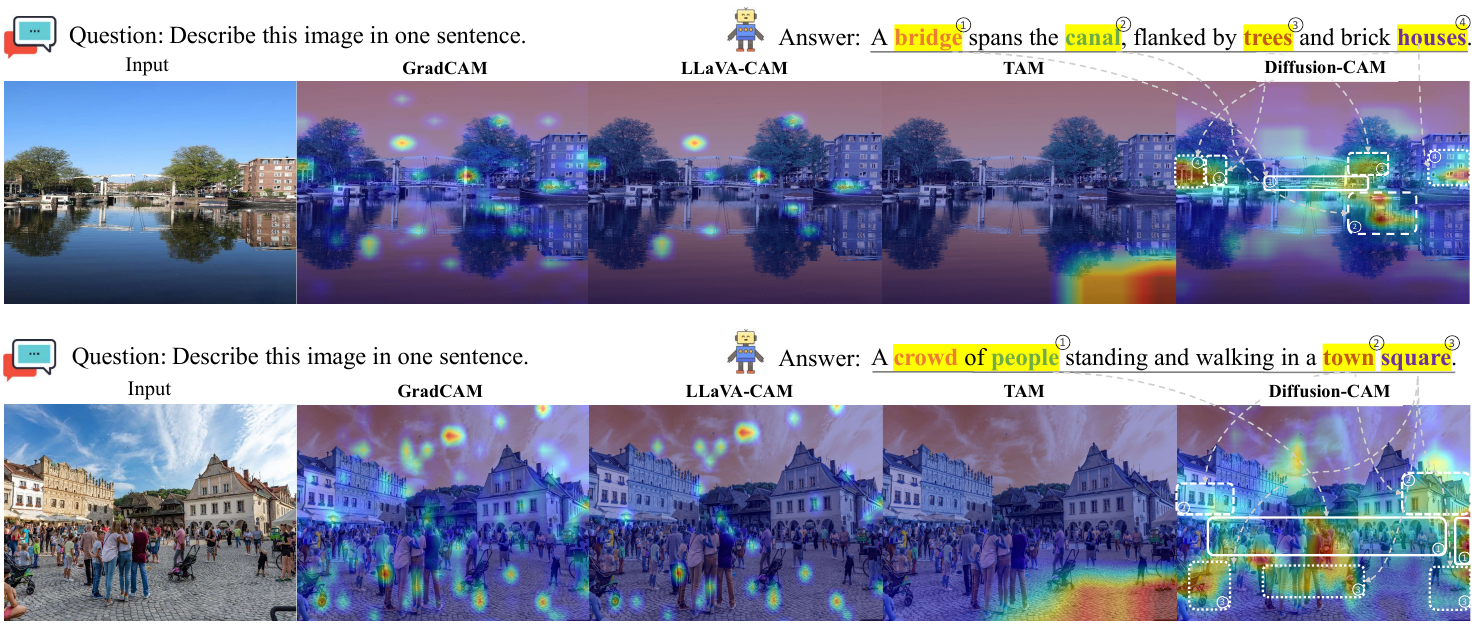}
	
	\caption{Visual comparison between our method and SoTA methods. Diffusion-CAM demonstrates superior global semantic alignment and also suppresses some noise in the baseline methods}
	\label{fig:token_selection_overall} 
\end{figure*}

\noindent \textbf{Localization Performance}
Figure~\ref{fig:layer_comparison} and Figure~\ref{fig:token_selection_overall} demonstrate the superior localization capability of our method across diverse scenarios. Diffusion-CAM produces more focused and clear activation maps that effectively highlight target objects while suppressing background noise.

Our approach consistently exhibits excellent localization performance across various contexts. In multi-object scenes, Diffusion-CAM successfully focuses on the most relevant regions while maintaining global context awareness. The activation maps display significantly reduced attention to uninformative background areas and substantially lower noise levels, with improved emphasis on target objects compared to baseline methods, demonstrate enhanced capability to distinguish between relevant and irrelevant regions.

\begin{figure}[htbp] 
	\centering
	\includegraphics[trim={1.5cm 7.5cm 2.4cm 2.3cm}, clip, width=1.01\linewidth]{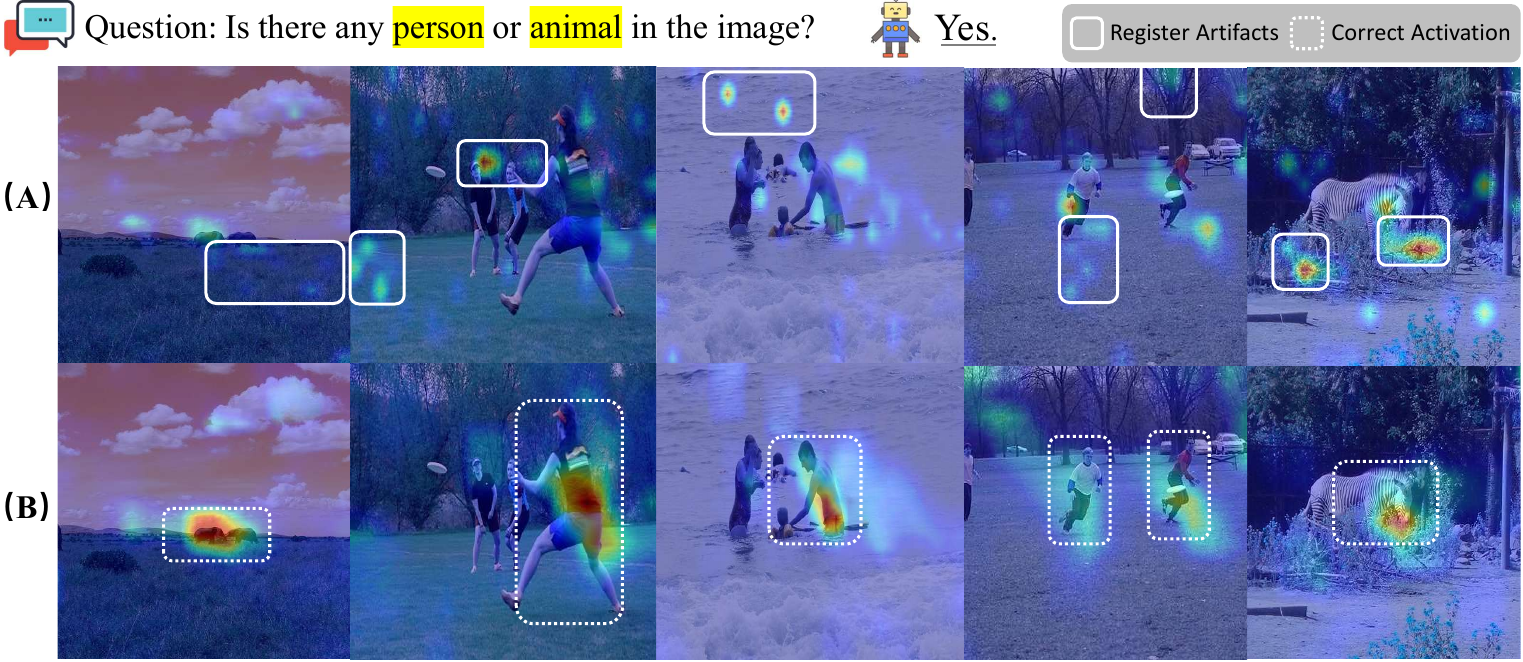}
	
	\caption{Visual comparison regarding artifacts elimination. (A) [Original CAM], (B) [Diffusion-CAM]. }
	\label{fig:token_selection_artifacts}
    
\end{figure}

\noindent \textbf{Visualization Comparison Across Target Layers}
Figure~\ref{fig:layer_comparison} presents CAM visualizations generated by our method across different target layers alongside comparative baseline results. Experimental findings indicate that our approach significantly outperforms baseline methods at all tested target layers. At layer[10], our method produces the most precise object localization, with activation regions more concentrated on target objects and minimal activation in irrelevant areas. 

As layer depth increases, activation patterns gradually become more diffuse because the image tokens receive substantially less attention than prompts and that image-token information flow converges in shallow layers and disperses in deep layers, with “cliff layers” after which image tokens become highly redundant.~\cite{zhang2024redundancy}while our method maintains effective target object localization. Crucially, even at deeper layers (layer[25]), it continues to generate clearer, more focused CAMs than baseline approaches, demonstrating robust adaptability to different layer features in diffusion models. This inter-layer consistency further validates that Diffusion-CAM effectively handles the unique global activation characteristics inherent to diffusion architectures.

\noindent\textbf{Suppression of Architectural Artifacts.} 
As illustrated in Figure \ref{fig:token_selection_artifacts}, the original activation maps without any processing frequently exhibit high-magnitude activations in irrelevant background regions, such as the sky or corners. These anomalies align with the ``register artifacts'' phenomenon inherent to Vision Transformers, where the model repurposes low-information background tokens to store global context~\cite{darcet2023vision}. These high-norm outliers manifests as false-positive noise in gradient-based explanations. In contrast, by incorporating our specialized modules, Diffusion-CAM effectively eliminates these high-frequency spikes. The resulting heatmaps demonstrate a clean background with activation concentrated solely on the semantic target, confirming our framework's robustness in mitigating the architectural artifacts of the underlying ViT encoder.

\noindent \textbf{Global Characteristics of Diffusion Models}
Figure~\ref{fig:token_selection_overall} illustrates the global characteristics of dMLLMs. The visualization reveals that, when describing landscapes or scenes without a specific target, Diffusion-CAM can more accurately focus on key tokens (such as nouns) in the response. Our method successfully enhances the quality of these global activations through targeted interventions, demonstrating the effectiveness of our global enhancement strategy.

\section{Conclusion}
We presented \textbf{Diffusion-CAM}, the first interpretability framework that bridges the explanatory gap for dMLLMs. By extracting features from the critical conditioning step and employing Adaptive Kernel Denoising, Single-Instance Causal Debiasing and other post-processing modules, our method showcases high-quality visualization results. Comprehensive evaluations confirm its superiority over autoregressive baselines in both localization accuracy and noise suppression. This work establishes a rigorous foundation for interpreting dMLLMs, paving the way for future more in-depth and comprehensive research.

\section{Limitations}
Despite the promising performance of Diffusion-CAM in interpreting dMLLMs, we acknowledge several limitations that point towards future research directions:

% \noindent\textbf{Architectural Specificity and Adaptation.} First, our current implementation is based on the LaViDa architecture, while the core principle of our framework—tracing gradients from the final denoising output back to the primary visual condition—is theoretically applicable to the broader class of dMLLMs, diverse architectural implementations pose practical tiny differences. Future work will focus on developing a unified interface to generalize Diffusion-CAM across heterogeneous diffusion architectures effortlessly.

\noindent\textbf{Abstract Concept Localization.} Second, while Diffusion-CAM demonstrates superior precision in grounding concrete physical objects, it exhibits limitations when interpreting abstract or emotional concepts. For prompts involving intangible attributes (e.g., love,'' guilt,'' or ``atmosphere''), the model often diffuses attention across the entire scene rather than converging on specific visual regions. This reflects an inherent \textit{spatial-semantic gap}: abstract concepts in MLLMs may rely on global holistic representations that lack distinct spatial coordinates, making them difficult to visualize through strictly gradient-based localization maps.
\bibliography{custom}

\appendix
\setcounter{equation}{0}
\renewcommand{\theequation}{\arabic{equation}}

\section{Background on CAM-based Interpretability}
\label{sec:appendix}

Class activation mapping (CAM)~\cite{zhou2016learning} is a post-hoc visual explanation technique that highlights the image regions most responsible for a target prediction. Let a classifier be denoted by $f:\mathcal{X}\rightarrow\mathbb{R}^{C}$, which maps an input image $\mathbf{x}\in\mathcal{X}$ to class logits $\mathbf{y}\in\mathbb{R}^{C}$. At a chosen layer, the network produces $K$ feature maps $\mathbf{A}=\{A_k\}_{k=1}^{K}$ with $A_k\in\mathbb{R}^{H\times W}$. CAM explains class $c$ by forming a weighted combination of these feature maps:
\begin{equation}
L^{c}=h\!\left(\sum_{k}\alpha_k^{c}A_k\right),
\end{equation}
where $\alpha_k^{c}$ measures the contribution of channel $k$ to class $c$, and $h(\cdot)$ is typically chosen as ReLU to retain positive evidence. In the original CAM formulation, when the classifier is built on global average pooling, the class logit can be written as
\begin{equation}
y^{c}=\sum_{k}w_k^{c}\,\mathrm{GAP}(A_k)+b^{c},
\end{equation}
so the channel importance is directly given by the classifier weight, i.e., $\alpha_k^{c}=w_k^{c}$~\cite{zhou2016learning}. Grad-CAM~\cite{selvaraju2017grad} generalizes this idea to broader architectures by replacing classifier weights with gradient-based importance:
\begin{equation}
\alpha_k^{c}=\frac{1}{Z}\sum_{i}\sum_{j}\frac{\partial y^{c}}{\partial A_{k}^{ij}},
\end{equation}
where $Z=H\times W$. Intuitively, CAM-style methods answer two coupled questions: \emph{what} target score should be explained, and \emph{how} intermediate feature channels should be weighted to recover spatial evidence.

Subsequent CAM variants mainly refine one of these two components. Score-CAM~\cite{wang2020score} replaces gradient-based weighting with forward confidence changes to reduce gradient noise; LayerCAM~\cite{jiang2021layercam} exploits spatially local importance from intermediate layers for finer localization; and Finer-CAM~\cite{zhang2025finer} further changes the explanation target from an isolated class score to a contrastive target, emphasizing discriminative details relative to visually similar concepts. More recently, multimodal explanation methods such as LLaVA-CAM~\cite{zhang2024redundancy} and Token Activation Map~\cite{li2025token} extend this CAM paradigm to large vision-language models by treating textual outputs or concepts as explanation targets, while Smooth-CAM-style designs~\cite{omeiza2019smooth} improve stability through perturbation-based smoothing. Taken together, these methods establish CAM as a lightweight and flexible framework for spatially grounded explanation.

\section{Shared Structural Basis of Diffusion MLLMs and Our CAM Adaptation}
Although recent diffusion MLLMs differ in emphasis, for example, LLaDA-V~\cite{you2025llada} adopts visual instruction tuning, Dream-V~\cite{ye2025dream2}L highlights planning-oriented diffusion backbones, and LaViDa-O~\cite{li2025lavidaO} / Sparse-LaViDa~\cite{li2025sparse} / LaViDa-R1~\cite{li2026lavida} extend the LaViDa~\cite{li2025lavida} family toward unified generation, sparse inference, and stronger reasoning, their underlying mechanism is highly consistent. In all these models, images are first converted into visual embeddings or visual tokens through a vision encoder and projector, then combined with textual prompts as \textbf{fixed multimodal conditioning}, while the response is generated by \textbf{iterative masked denoising}. In other words, despite architectural variations, they all expose the same attribution interface: intermediate multimodal hidden states that support \textbf{image-conditioned prediction of masked response tokens}. This shared conditional masked-diffusion structure is the structural basis targeted by our method.
\label{app:shared_dmlm_cam}
\begin{table}[htbp]
\centering
\small
\renewcommand{\arraystretch}{1.4} % 保持舒适的行高
\setlength{\tabcolsep}{4pt}      
\begin{tabular}{lccc}
\toprule
Attribution step & feat\_len & img\_end & Valid \\
\midrule
$t=0$ (Prefix) & 579 & 551 & PASS \\
$t=S/4$ & 64 & 551 & FAIL \\
$t=S/2$ & 64 & 551 & FAIL \\
$t=3S/4$ & 64 & 551 & FAIL \\
Aggregate over steps & -- & -- & N/A \\
\bottomrule
\end{tabular}
\vspace{0.3em}
\caption{Per-step feasibility check for image-span CAM extraction under LaViDa's Prefix-DLM + KV-cached generation. A denoising step is considered valid only if the hooked hidden-state length satisfies $\texttt{seq\_len} \ge \texttt{img\_end}$. Across all logged step-records ($6600 = 200$ images $\times$ $(1$ prefix $+ S$ denoising steps$)$), only the earliest conditioning step passes, yielding a valid ratio of $200/6600 = 3.0\%$.}
\label{tab:feasibility_check}
\end{table}

\begin{table*}[!t]
\centering
\small
\renewcommand{\arraystretch}{1.4} % 保持舒适的行高
\setlength{\tabcolsep}{4pt}      
\begin{tabular}{lccccc}
\toprule
Attribution step & Valid & Contrast $\uparrow$ & Concen. $\uparrow$ & Obj-IoU $\uparrow$ & F3-score $\uparrow$ \\
\midrule
Prefix ($t=0$) & 200/200 & $2.549 \pm 0.126$ & $50.69 \pm 1.741$ & $0.307 \pm 0.024$ & $0.234 \pm 0.018$ \\
$t=S/4$ & 0/200 & -- & -- & -- & -- \\
$t=S/2$ & 0/200 & -- & -- & -- & -- \\
$t=3S/4$ & 0/200 & -- & -- & -- & -- \\
Aggregate over steps & -- & -- & -- & -- & -- \\
\bottomrule
\end{tabular}
\caption{CAM quality at the valid attribution step. Since later denoising steps are infeasible for image-span CAM extraction under Prefix-DLM + KV cache, metrics are reported only for the valid prefix step.}
\label{tab:valid_step_quality}
\end{table*}
To make the model-aware extraction rule explicit, we report the per-step feasibility check and the CAM quality at the valid step under LaViDa's~\cite{li2025lavida} Prefix-DLM + KV-cached generation. As shown in Table~\ref{tab:feasibility_check}, under Prefix-DLM + KV-cached generation only the earliest conditioning step remains structurally valid for image-span CAM extraction. We therefore report CAM quality only at this valid step in Table~\ref{tab:valid_step_quality}.

This rule explains why LaViDa~\cite{li2025lavida} typically yields only the earliest conditioning step as valid under Prefix-DLM + KV cache, while models that expose the full multimodal sequence at more denoising steps can admit multiple valid extraction steps. Therefore, our method is applicable to diffusion MLLMs in general: it relies only on the common masked-denoising mechanism and on the availability of at least one structurally valid intermediate multimodal state, both of which are shared across current dMLLM families.

\section{Additional Technical Analyses}
\label{app:additional_analyses}

\subsection{Sensitivity Analysis of Distribution-Aware Confidence Gating}
\label{app:dacg_sensitivity}

The proposed Distribution-Aware Confidence Gating (DACG) introduces several threshold-like parameters, including the branch-boundary values and branch-specific percentile settings. To evaluate whether the method is overly sensitive to these choices, we perform two targeted sensitivity studies on 200 COCO images with the prompt \textit{``Describe this image in one sentence.''} We report the same metrics as in the main paper, namely Contrast, Concentration, Obj-IoU, and F3-score, together with the branch routing ratios.

\paragraph{Branch-boundary sensitivity.}
We first vary the three branch-boundary parameters one at a time while keeping the remaining settings fixed at their default values. Specifically, the baseline setting is $\delta_\sigma=0.22$, $\delta_\mu=0.35$, and $\delta'_\mu=0.25$, with branch percentiles $\alpha_1=90$, $\alpha_2=75$, $\alpha_3=80$, and $\alpha_4=85$. Tables~\ref{tab:dacg_sigma}, \ref{tab:dacg_mu}, and \ref{tab:dacg_mup} report the detailed sweep results. Across all three sweeps, the performance varies smoothly without catastrophic degradation. For example, Obj-IoU changes within a narrow range for each sweep ($0.201$--$0.215$ for $\delta_\sigma$, $0.199$--$0.215$ for $\delta_\mu$, and $0.198$--$0.215$ for $\delta'_\mu$), while F3-score remains within $0.192$--$0.196$ throughout. We also observe that the branch routing ratios shift gradually and always sum to approximately $100\%$, confirming that each sample is deterministically assigned to exactly one branch. These results indicate that DACG is not brittle with respect to modest perturbations of its boundary parameters.

\begin{table*}[t]
\centering
\small
\renewcommand{\arraystretch}{1.4} % 保持舒适的行高
\setlength{\tabcolsep}{4pt}      
\resizebox{\textwidth}{!}{
\begin{tabular}{c c c c c c c c c}
\toprule
$\delta_\sigma$ & Obj-IoU $\uparrow$ & Contrast $\uparrow$ & Concen. $\uparrow$ & F3-Score $\uparrow$ & high\_var & high\_mean & low\_mean & default \\
\midrule
0.18 & $0.201 \pm 0.021$ & $2.29 \pm 0.13$ & $41.89 \pm 2.044$ & $0.192 \pm 0.013$ & 36.4\% & 14.1\% & 47.2\% & 3.3\% \\
0.20 & $0.208 \pm 0.017$ & $2.38 \pm 0.08$ & $42.77 \pm 2.041$ & $0.194 \pm 0.012$ & 27.8\% & 16.6\% & 50.8\% & 4.8\% \\
\textbf{0.22} & $\mathbf{0.215 \pm 0.015}$ & $\mathbf{2.44 \pm 0.12}$ & $\mathbf{43.21 \pm 2.037}$ & $\mathbf{0.196 \pm 0.011}$ & 22.1\% & 19.3\% & 53.4\% & 5.2\% \\
0.24 & $0.206 \pm 0.019$ & $2.39 \pm 0.12$ & $42.69 \pm 2.039$ & $0.193 \pm 0.015$ & 18.6\% & 21.0\% & 54.9\% & 5.6\% \\
0.26 & $0.202 \pm 0.025$ & $2.30 \pm 0.14$ & $42.01 \pm 2.049$ & $0.192 \pm 0.011$ & 15.1\% & 22.3\% & 56.4\% & 6.2\% \\
\bottomrule
\end{tabular}
}
\caption{$\delta_\sigma$ sweep for DACG branch-boundary sensitivity. (Baseline values are $\delta_\sigma=0.22$, $\delta_\mu=0.35$, $\delta'_\mu=0.25$, with $\alpha_1=90$, $\alpha_2=75$, $\alpha_3=80$, and $\alpha_4=85$.)}
\label{tab:dacg_sigma}
\end{table*}

\begin{table*}[t]
\centering
\small
\renewcommand{\arraystretch}{1.4} % 保持舒适的行高
\setlength{\tabcolsep}{4pt}      
\resizebox{\textwidth}{!}{
\begin{tabular}{c c c c c c c c c}
\toprule
$\delta_\mu$ & Obj-IoU $\uparrow$ & Contrast $\uparrow$ & Concen. $\uparrow$ & F3-Score $\uparrow$ & high\_var & high\_mean & low\_mean & default \\
\midrule
0.28 & $0.199 \pm 0.024$ & $2.30 \pm 0.15$ & $41.75 \pm 2.121$ & $0.192 \pm 0.037$ & 14.7\% & 37.2\% & 45.2\% & 2.9\% \\
0.32 & $0.205 \pm 0.022$ & $2.36 \pm 0.18$ & $42.41 \pm 2.132$ & $0.194 \pm 0.034$ & 19.2\% & 26.4\% & 49.8\% & 4.6\% \\
\textbf{0.35} & $\mathbf{0.215 \pm 0.015}$ & $\mathbf{2.44 \pm 0.12}$ & $\mathbf{43.21 \pm 2.037}$ & $\mathbf{0.196 \pm 0.011}$ & 22.1\% & 19.3\% & 53.4\% & 5.2\% \\
0.38 & $0.207 \pm 0.027$ & $2.35 \pm 0.19$ & $42.60 \pm 2.054$ & $0.194 \pm 0.031$ & 25.6\% & 11.8\% & 57.2\% & 5.4\% \\
0.42 & $0.201 \pm 0.025$ & $2.32 \pm 0.19$ & $42.01 \pm 2.173$ & $0.193 \pm 0.039$ & 29.3\% & 3.1\% & 61.7\% & 5.9\% \\
\bottomrule
\end{tabular}
}
\caption{$\delta_\mu$ sweep for DACG branch-boundary sensitivity.}
\label{tab:dacg_mu}
\end{table*}

\begin{table*}[!t]
\centering
\small
\renewcommand{\arraystretch}{1.4} % 保持舒适的行高
\setlength{\tabcolsep}{4pt}      
\resizebox{\textwidth}{!}{
\begin{tabular}{c c c c c c c c c}
\toprule
$\delta'_\mu$ & Obj-IoU $\uparrow$ & Contrast $\uparrow$ & Concen. $\uparrow$ & F3-Score $\uparrow$ & high\_var & high\_mean & low\_mean & default \\
\midrule
0.20 & $0.198 \pm 0.025$ & $2.28 \pm 0.22$ & $40.81 \pm 2.201$ & $0.192 \pm 0.012$ & 26.0\% & 23.5\% & 43.6\% & 6.9\% \\
0.23 & $0.204 \pm 0.023$ & $2.31 \pm 0.19$ & $42.07 \pm 2.147$ & $0.194 \pm 0.013$ & 23.4\% & 20.7\% & 49.9\% & 6.0\% \\
\textbf{0.25} & $\mathbf{0.215 \pm 0.015}$ & $\mathbf{2.44 \pm 0.12}$ & $\mathbf{43.21 \pm 2.037}$ & $\mathbf{0.196 \pm 0.011}$ & 22.1\% & 19.3\% & 53.4\% & 5.2\% \\
0.28 & $0.209 \pm 0.021$ & $2.39 \pm 0.17$ & $42.61 \pm 2.094$ & $0.195 \pm 0.015$ & 19.2\% & 16.8\% & 60.2\% & 3.8\% \\
0.30 & $0.202 \pm 0.031$ & $2.25 \pm 0.24$ & $41.98 \pm 2.188$ & $0.193 \pm 0.013$ & 14.9\% & 13.4\% & 69.3\% & 2.4\% \\
\bottomrule
\end{tabular}
}
\caption{$\delta'_\mu$ sweep for DACG branch-boundary sensitivity.}
\label{tab:dacg_mup}
\end{table*}

\begin{table*}[!t]
\centering
\small
\renewcommand{\arraystretch}{1.5} % 保持舒适的行高
\setlength{\tabcolsep}{4pt}      
\resizebox{\textwidth}{!}{
\begin{tabular}{l c c c c c c c c}
\toprule
Setting & Obj-IoU $\uparrow$ & Contrast $\uparrow$ & Concen. $\uparrow$ & F3-Score $\uparrow$ & high\_var & high\_mean & low\_mean & default \\
\midrule
\textbf{Dynamic (ours)} & $\mathbf{0.215 \pm 0.015}$ & $\mathbf{2.44 \pm 0.12}$ & $\mathbf{43.21 \pm 2.037}$ & $\mathbf{0.196 \pm 0.011}$ & 22.1\% & 19.3\% & 53.4\% & 5.2\% \\
All $\alpha -5$ & $0.201 \pm 0.022$ & $2.28 \pm 0.26$ & $41.77 \pm 2.111$ & $0.193 \pm 0.008$ & 22.1\% & 19.3\% & 53.4\% & 5.2\% \\
All $\alpha +5$ & $0.203 \pm 0.017$ & $2.32 \pm 0.28$ & $41.62 \pm 2.102$ & $0.193 \pm 0.013$ & 22.1\% & 19.3\% & 53.4\% & 5.2\% \\
Fixed $\alpha = 85$ & $0.197 \pm 0.019$ & $2.23 \pm 0.23$ & $41.05 \pm 1.966$ & $0.192 \pm 0.009$ & 22.1\% & 19.3\% & 53.4\% & 5.2\% \\
\bottomrule
\end{tabular}
}
\caption{Percentile sensitivity of DACG with branch boundaries fixed.}
\label{tab:dacg_alpha}
\end{table*}

\begin{figure*}[!t] 
	\centering
	\includegraphics[trim={0.5cm 5.5cm 1cm 5.2cm}, clip, width=1.01\linewidth]{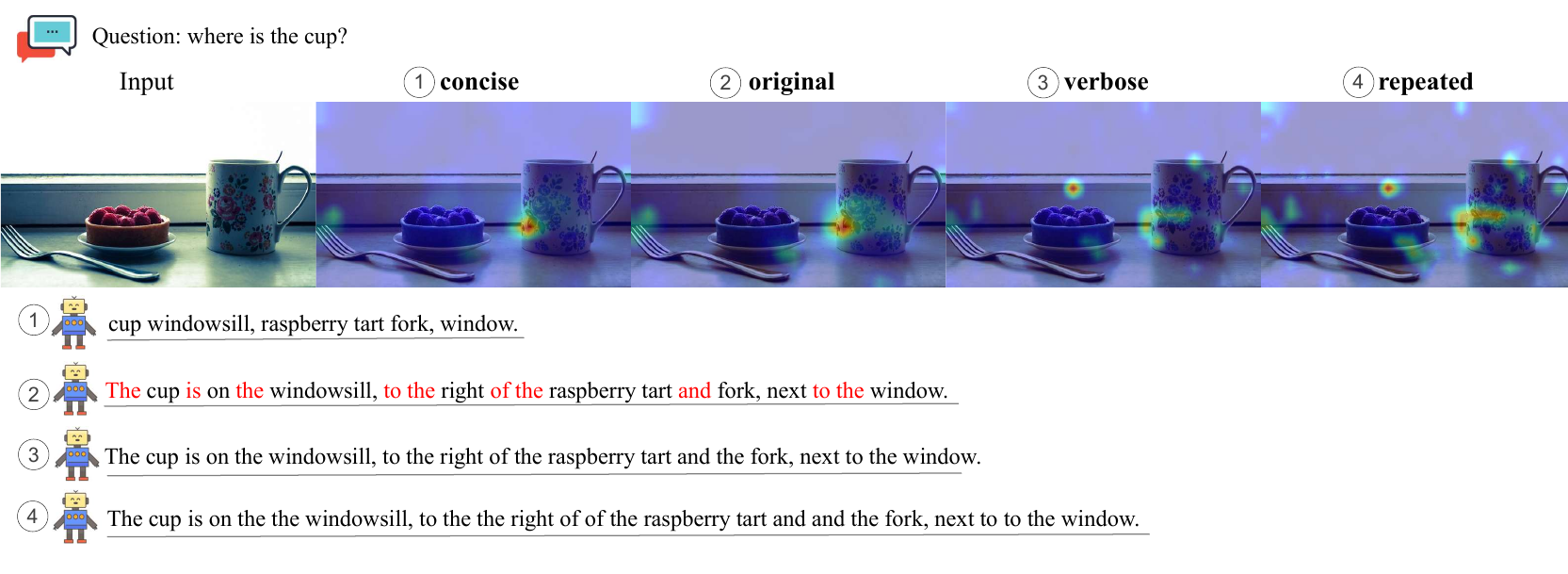}
	\caption{Visualization results of four caption variants under the same teacher-forced attribution protocol by Diffusion-CAM. It is evident that, following the addition of redundant function words, regions in the image unrelated to the answer keywords (e.g., "cup") began to exhibit high-intensity activation; furthermore, as the number of function words increased, these irrelevant activations became increasingly pronounced. }
	\label{fig:SICD}
\end{figure*}

\paragraph{Percentile sensitivity.}
We then evaluate the sensitivity to the percentile parameters while fixing the branch boundaries to their baseline values. Table~\ref{tab:dacg_alpha} summarizes the results. The percentile choices are likewise stable: Obj-IoU varies only from $0.197$ to $0.215$, and F3-score remains within $0.192$--$0.196$ across the tested configurations. Importantly, in this setting the branch routing ratios remain unchanged because the partition boundaries are fixed; only the within-branch processing strength changes. This further supports our claim that DACG acts as a mild confidence-aware refinement rather than a brittle thresholding mechanism. Overall, the sensitivity study shows that DACG improves the baseline in a stable manner, and that its effectiveness does not hinge on a single narrowly tuned hyperparameter choice.

\begin{table}[htbp]
\centering
\small
\renewcommand{\arraystretch}{1.4} % 保持舒适的行高
\setlength{\tabcolsep}{4pt}      
\begin{tabular}{@{}lccc@{}}
\toprule
Stage & Time (ms) & Device & Complexity \\
\midrule
Base attribution pass & 4880.8 & GPU & - \\
AKD & 12.7 & CPU & $O(HW \cdot k^2)$ \\
DACG & 6.9 & CPU & $O(HW)$ \\
CBA & 5.6 & CPU & $O(HW)$ \\
SICD & 10.2 & CPU & $O(HW)$ \\
\textbf{Total post-processing} & \textbf{35.4} & CPU & $O(HW)$ \\
\bottomrule
\end{tabular}
\caption{Per-module runtime breakdown. The refinement modules are CPU-only and add negligible overhead relative to the attribution pass.}
\label{tab:overhead_breakdown}
\end{table}

\subsection{Efficiency and Computational Overhead}
\label{app:efficiency}
\begin{table*}[!t]
\centering
\small
\renewcommand{\arraystretch}{1.4} % 保持舒适的行高
\setlength{\tabcolsep}{7pt}        % 稍微增加列宽，让数字呼吸
\begin{tabular}{lcccc}
\toprule
Method & Time (ms) & Peak Mem (GB) & F3-score & Notes \\
\midrule
Diffusion-CAM (base) & 4880.8 & 83.21 & 0.188 & generate + contrastive backward \\
Base + AKD & 4893.8 & 83.21 & 0.207 & + rank Gaussian filter \\
Base + DACG & 4887.0 & 83.21 & 0.187 & + confidence gating \\
Base + CBA & 4886.3 & 83.21 & 0.190 & + background attenuation \\
Base + SICD & 4891.1 & 83.21 & 0.195 & + causal debiasing \\
Diffusion-CAM (full) & 4914.4 & 83.21 & 0.225 & all modules enabled \\
\bottomrule
\end{tabular}
\caption{End-to-end computational overhead per image on 200 images (NVIDIA H20).}
\label{tab:overhead_e2e}
\end{table*}

\begin{table*}[!t]
\centering
\small
\renewcommand{\arraystretch}{1.5} % 保持舒适的行高
\setlength{\tabcolsep}{10pt}        % 稍微增加列宽，让数字呼吸
\begin{tabular}{lcccc}
\toprule
Variant & Contrast $\uparrow$ & Concen. $\uparrow$ & Obj-IoU $\uparrow$ & F3-score $\uparrow$ \\
\midrule
concise & $2.3546 \pm 0.1133$ & $43.56 \pm 1.968$ & $0.2342 \pm 0.0153$ & $0.1942 \pm 0.0071$ \\
original & $2.1905 \pm 0.1219$ & $41.31 \pm 2.029$ & $0.2086 \pm 0.0166$ & $0.1846 \pm 0.0085$ \\
verbose & $2.1266 \pm 0.1286$ & $39.17 \pm 2.031$ & $0.1969 \pm 0.0155$ & $0.1829 \pm 0.0087$ \\
repeated & $2.0985 \pm 0.1300$ & $39.01 \pm 2.027$ & $0.1908 \pm 0.0138$ & $0.1811 \pm 0.0092$ \\
\bottomrule
\end{tabular}
\caption{Controlled validation of the linguistic economy hypothesis on 200 COCO images. We compare four caption variants under the same teacher-forced attribution protocol. Increasing function-word redundancy leads to more diffuse and less object-specific CAMs.}
\label{tab:linguistic_economy}
\end{table*}
Since our framework contains multiple post-processing modules, an important practical question is whether these refinements substantially increase computational overhead. To answer this, we measure both end-to-end latency and per-module runtime on 200 images using an NVIDIA H20 GPU. We also report the peak GPU memory footprint. The key observation is that the dominant cost comes from the attribution pass itself, i.e., generation followed by the contrastive backward pass required for gradient-based CAM extraction. In contrast, all refinement modules operate only on the final 2D activation map and simple map statistics, and therefore do not introduce any additional model forward or backward passes. 
Table~\ref{tab:overhead_e2e} reports the end-to-end runtime of the base Diffusion-CAM and its module variants, while Table~\ref{tab:overhead_breakdown} provides a stage-wise breakdown. 

Quantitatively, the full pipeline requires $4914.4$ ms per image, compared with $4880.8$ ms for the base attribution pass, which means that all post-processing modules together add only $35.4$ ms, i.e., \textbf{0.7\%} of the total runtime. Moreover, the peak GPU memory remains unchanged at $83.21$ GB across all variants, since the additional modules run on CPU and do not increase the memory footprint of the model itself. The per-module breakdown further shows that each individual module accounts for only a small fraction of total time. These results indicate that the proposed refinements provide improved explanation quality at negligible extra cost beyond the original attribution pass. We therefore view the framework as a practical interpretability tool whose main computational bottleneck remains the gradient-based CAM extraction itself, rather than the subsequent refinement modules.

\section{Linguistic Economy Hypothesis and Controlled Validation}
\label{app:linguistic_economy}

Our Single-Instance Causal Debiasing module is motivated by the observation that redundant syntactic scaffolding and repeated function words often induce diffuse, non-specific activations in multimodal heatmaps. We emphasize that, in this paper, we do not use ``linguistic economy'' as a universal linguistic claim. Instead, we adopt it as an empirical hypothesis about model explanations: when semantically equivalent captions are expressed with increasingly redundant function-word structure, the resulting CAMs tend to become less concentrated and less object-specific.

To test this hypothesis in a controlled manner, we conduct a teacher-forced language intervention study on 200 COCO images. For each image, we construct four caption variants derived from the same base description: \textit{concise} (content words only), \textit{original}, \textit{verbose} (with additional function-word scaffolding), and \textit{repeated} (where function words are duplicated once). We then run the same hooked activation-and-gradient attribution pipeline as in our main method, keeping the image, hook layer, target-score construction, and evaluation metrics fixed across all variants. 

Table~\ref{tab:linguistic_economy} reports the results: moving from concise to original to verbose to repeated captions consistently reduces Contrast, Concentration, Obj-IoU, and F3-score. In particular, as shown in Figure~\ref{fig:SICD}, repeated functional words lead to the most diffuse and least localized heatmaps. This result provides direct evidence that function-word redundancy can act as an interference source in diffusion-CAM attribution, and therefore supports the design motivation behind the Single-Instance Causal Debiasing module. In summary, this experiment does not attempt to prove a general theory of language production. Rather, it verifies, within our visual explanation setting, that syntactic redundancy is systematically associated with weaker spatial grounding and more diffuse activation patterns.

\end{document}